\documentclass[11pt,a4paper]{article}
\usepackage{times,latexsym}
\usepackage{url}
\usepackage[T1]{fontenc}
\usepackage[table]{xcolor}
\usepackage[acceptedWithA]{tacl2021v1}

\usepackage{xspace,mfirstuc,tabulary}

\newif\iftaclinstructions
\taclinstructionsfalse 
\iftaclinstructions
\renewcommand{\confidential}{}
\renewcommand{\anonsubtext}{(No author info supplied here, for consistency with
TACL-submission anonymization requirements)}
\newcommand{\instr}
\fi

\iftaclpubformat 

\else

\fi

\usepackage{microtype}

\usepackage{inconsolata}

\usepackage[pdftex]{graphicx}
\usepackage{makecell}
\usepackage{enumitem}
\usepackage{booktabs}
\usepackage{arydshln}
\usepackage{amsmath}
\usepackage{multirow}
\usepackage{nicematrix}
\usepackage[capitalise]{cleveref}
\usepackage{pifont}
\usepackage{fontawesome}
\usepackage{amssymb}
\usepackage{textcomp}
\usepackage{amsfonts} 
\usepackage{soul}
\usepackage{CJKutf8}
\crefname{section}{\S}{\S\S}
\Crefname{section}{\S}{\S\S}
\crefname{appendix}{Appendix}{Appendices}
\Crefname{appendix}{Appendix}{Appendices}

\definecolor{Melon}{RGB}{241, 162, 130}
\sethlcolor{Melon}
\DeclareMathSymbol{\shortminus}{\mathbin}{AMSa}{"39}
\newcommand{\ii}[1]{\ensuremath{\textcolor{teal}{_{\bf \uparrow#1}}}}
\newcommand{\dd}[1]{\ensuremath{\textcolor{purple}{_{\bf \downarrow#1}}}}
\newcommand{\DD}[1]{\ensuremath{\textcolor{purple}{{\downarrow#1}}}}
\newcommand{\nn}[1]{\ensuremath{\textcolor{gray}{_{\bf \shortminus#1}}}}

\newcommand{\zh}[1]{\begin{CJK*}{UTF8}{gkai}#1\end{CJK*}}

\definecolor{xing}{RGB}{18, 21, 141}

\definecolor{hzw}{RGB}{223, 97, 76}

\title{Exploring Human-Like Translation Strategy \\with Large Language Models}

\author{%
    \thanks{\quad Zhiwei and Tian contributed equally and are co-first authors. Work was done when Zhiwei and Tian were interning at Tencent AI Lab.}$\,$ Zhiwei He$^1$\quad $^*$Tian Liang$^2$\quad Wenxiang Jiao$^3$\quad Zhuosheng Zhang$^1$\\
    \bf Yujiu Yang$^2$\quad \thanks{\quad Rui Wang and Zhaopeng Tu are co‐corresponding authors.}$\,\,$Rui Wang$^1$\quad $\,$$^\dagger$Zhaopeng Tu$^3$\quad Shuming Shi$^3$\quad Xing Wang$^3$\\
    $^1$Shanghai Jiao Tong University\ \ \ $^2$Tsinghua University\ \ \ $^3$Tencent AI Lab\\
    $^1$\texttt{\small{\{zwhe.cs,zhangzs,wangrui12\}}@sjtu.edu.cn} \\
    $^2$\texttt{\small{\{liangt21@mails,yang.yujiu@sz\}}.tsinghua.edu.cn} \\
    $^3$\texttt{\small{\{joelwxjiao,shumingshi,zptu,brightxwang\}}@tencent.com}
}
\date{}

\begin{document}
\maketitle
\begin{abstract}
Large language models (LLMs) have demonstrated impressive capabilities in general scenarios, exhibiting a level of aptitude that approaches, in some aspects even surpasses, human-level intelligence.
Among their numerous skills, the translation abilities of LLMs have received considerable attention.
Compared to typical machine translation that focuses solely on source-to-target mapping, LLM-based translation can potentially mimic the human translation process which might take preparatory steps to ensure high-quality translation.
This work explores this possibility by proposing the \textbf{MAPS} framework, which stands for \textbf{M}ulti-\textbf{A}spect \textbf{P}rompting and \textbf{S}election.
Specifically, we enable LLMs first to analyze the given source sentence and induce three aspects of translation-related knowledge: keywords, topics, and relevant demonstrations to guide the final translation process. Moreover, we employ a selection mechanism based on quality estimation to filter out noisy and unhelpful knowledge.
Both automatic (3 LLMs $\times$ 11 directions $\times$ 2 automatic metrics) and human evaluation (preference study and MQM) demonstrate the effectiveness of MAPS.
Further analysis shows that by mimicking the human translation process, MAPS reduces various translation errors such as hallucination, ambiguity, mistranslation, awkward style, untranslated text, and omission.
Source code is available at \url{https://github.com/zwhe99/MAPS-mt}.
\end{abstract}

\section{Introduction}

Large language models (LLMs) have recently demonstrated remarkable general capabilities across a wide range of tasks, making substantial strides in the field of artificial general intelligence (AGI).
These capabilities have led to LLMs exhibiting a certain degree of human-level intelligence, particularly in the areas of language understanding and generation~\cite{liang2022holistic,bubeck2023sparks,Wu2023ChatGPTOG,moghaddam2023boosting}.
Among the numerous tasks, translation has emerged as a prominent area where LLMs have shown impressive capacity and competence~\cite{jiao2023ischatgpt,agrawal-etal-2023-context,zhang2023prompting,vilar2022prompting,moslem2023adaptive,pilault2023interactive,garcia2023unreasonable,hendy2023good,zhu2023multilingual,jiao2023parrot,wang2023document,karpinska2023large,Peng2023ChatGPT4MT,lyu2023new,bawden2023investigating,lu2023chain}.
This progress above harkens back to the long-term aspirations and dreams of earlier machine translation research in the 1960s~\cite{bar1960demonstration, macklovitch1995future}: Can LLMs employ a translation process similar to human translators?

\begin{figure*}[t]
    \centering
    \includegraphics[width=0.8\linewidth]{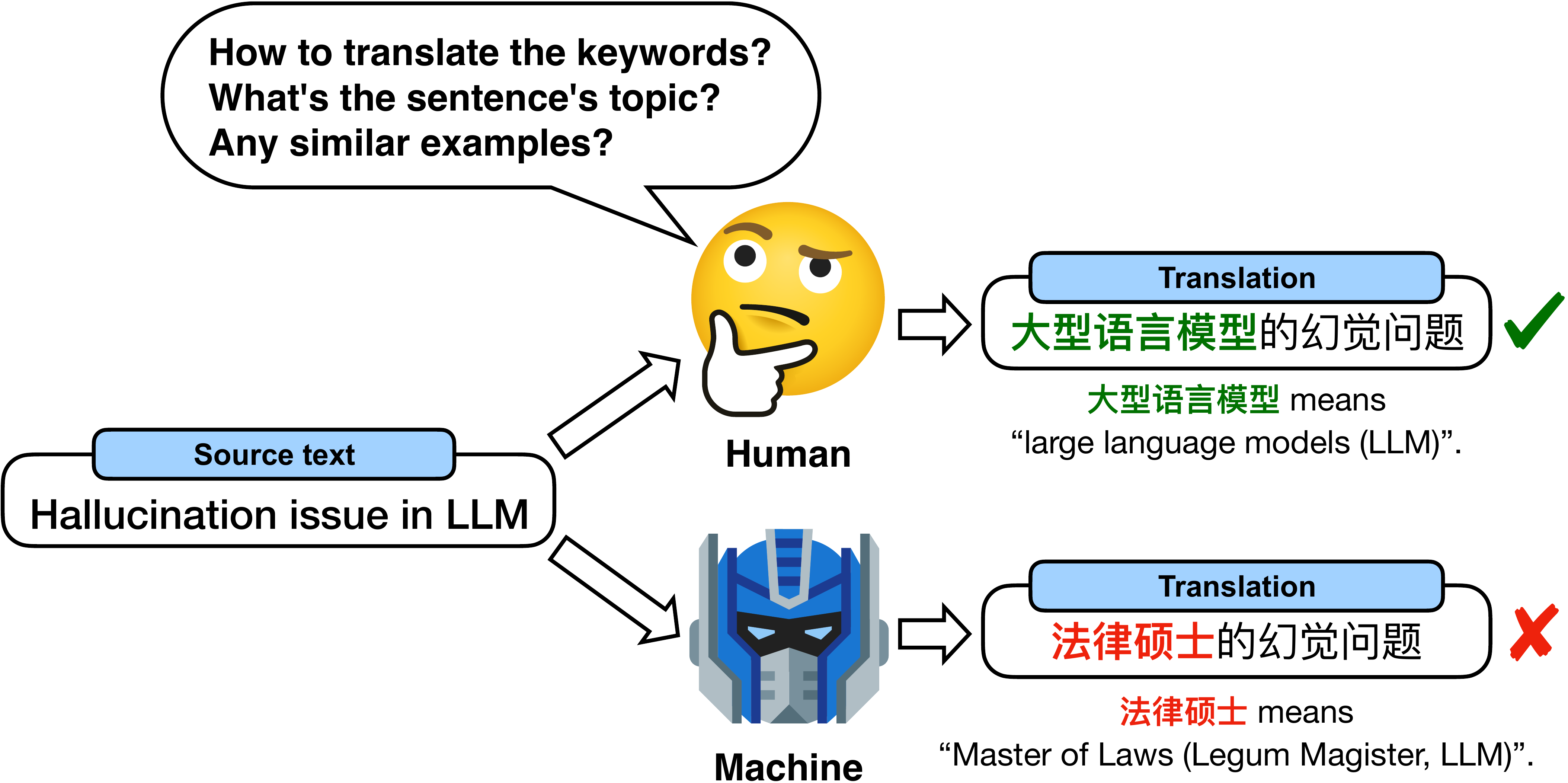}
    \caption{The difference between machine and human translation in an English$\rightarrow$Chinese example. Typical neural machine translation is a source-to-target mapping process, while human translators can take complex steps to ensure the quality and accuracy of the translation.}
    \label{fig:intro}
\end{figure*}

\figurename~\ref{fig:intro} illustrates the difference between the processes of machine and human translation.
While conventional machine translation is typically a direct source-to-target mapping process, professional human translators tend to take preparatory steps when working with the given source text, including gathering and meticulously analyzing information such as keywords, topics, and relevant example sentences~\cite{baker2018other,koehn2009process,bowker2002computer,hatim2004translation}.
These steps are critical for ensuring high-quality translations that accurately capture the nuances of the source material.
Although recent advances in LLM research indicate that current LLMs are approaching human-like general intelligence~\cite{bubeck2023sparks,park2023generative}, the extent to which LLMs can emulate such strategies remains underexplored.

The primary focus of this paper is to explore whether LLMs can imitate the translation strategies employed by human translators.
Specifically, we aim to investigate whether LLMs can effectively preprocess the source text and leverage the relevant knowledge to improve their translations.

To this end, we propose a method called \textbf{MAPS}, which stands for \textbf{M}ulti-\textbf{A}spect \textbf{P}rompting and \textbf{S}election.
MAPS prompts the LLMs to analyze the source sentence and elicit translation-related knowledge in three aspects: \textbf{keywords}, \textbf{topics}, and \textbf{relevant demonstrations}.
This knowledge then guides the LLM toward generating more accurate translations.
To further enhance translation quality, we also employ a post-selection process to filter out unhelpful knowledge and select the best translation based on reference-free quality estimation (QE).
We validate our approach across 11 translation directions (covering high-, medium- and low-resource language pairs) from WMT22~\cite{kocmi-etal-2022-findings} and 3 LLMs (text-davinci-003, Alpaca and Vicuna).
Automatic evaluation shows that MAPS achieves significant improvement over other baselines in terms of COMET and BLEURT.
Further analysis emphasizes the importance of the extracted knowledge in resolving hallucination and ambiguity in translation.
We also conduct human preference studies and Multidimensional Quality Metrics (MQM) evaluation~\cite{burchardt-2013-multidimensional} which show that MAPS produces more favorable translations by reducing mistranslation, awkward style, untranslated text, and omission errors.

In contrast to other LLM-based translation approaches, such as Dictionary-based Prompting~\cite{ghazvininejad2023dictionary} and In-context Learning (ICL)~\cite{agrawal-etal-2023-context}, MAPS focuses on translating general scenarios without any preconceived assumptions about the domain of translation.
As a result, MAPS does not require the preparation of any external ``datastore'', which might include a meticulously constructed glossary~\cite{moslem2023adaptive}, dictionary~\cite{ghazvininejad2023dictionary}, or sample pool~\cite{agrawal-etal-2023-context}, for specific language pairs and domains in advance.

In summary, the contributions of this work are detailed as follows:
\begin{itemize}[leftmargin=10pt]
    \item Inspired by human translation strategy, we propose the MAPS method, which mimics the human process of analyzing the source text to gather useful knowledge, ultimately leading to an accurate translation.

    \item We demonstrate that the three types of translation-related knowledge (keywords, topics, and relevant demonstrations) complement each other.
    The best translation performance can be achieved by using all three types of knowledge simultaneously.

    \item Our in-depth analyses of MAPS, encompassing both automatic and human evaluations, demonstrates its proficiency in resolving ambiguities and reducing hallucinations and other prevalent translation errors.
    Furthermore, we examined the inference time of MAPS and investigated potential acceleration techniques.
\end{itemize}
\begin{figure*}[htbp!]
    \centering
    \includegraphics[width=\linewidth]{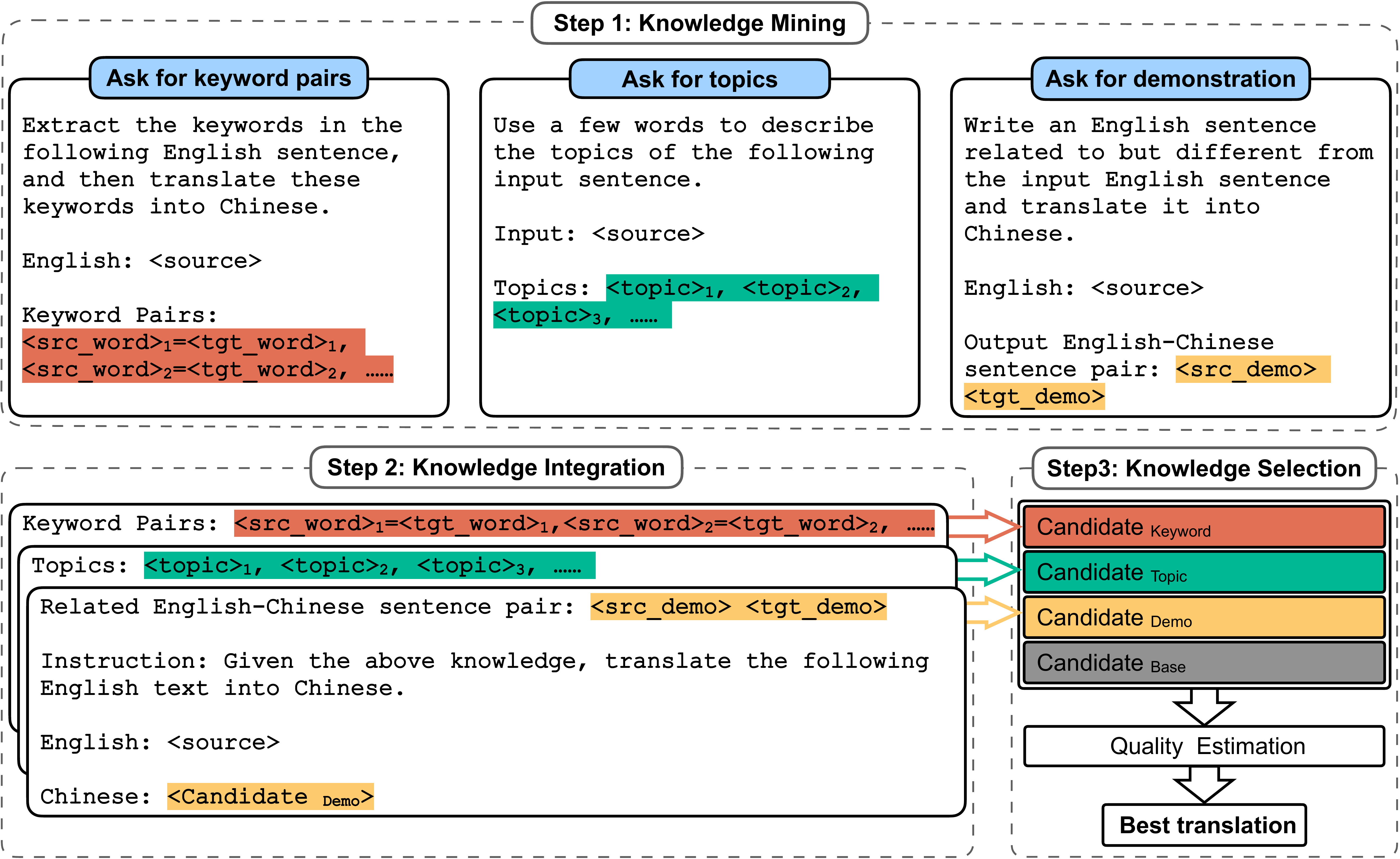}
    \caption{Framework of MAPS. On a high level, MAPS consists of three stages: \textit{(1) Knowledge Mining}: the LLM analyzes the source sentence and generates three aspects of knowledge useful for translation: keywords, topics, and relevant demonstration; \textit{(2) Knowledge Integration}: guided by different types of knowledge separately, the LLM generates multiple translation candidates; \textit{(3) Knowledge Selection}: the candidate deemed best by the QE is selected as the final translation. Best viewed in color.}
    \label{fig:method}
\end{figure*}
\section{MAPS: Multi-Aspect Prompting and Selection}
\label{sec:MAPS}

In this section, we introduce the MAPS framework.
As depicted in \figurename~\ref{fig:method}, MAPS consists of three steps --- knowledge mining, integration, and selection.
When mining the knowledge, the LLM operates in a manner akin to a human translator, analyzing the source text and generating background knowledge that is beneficial to translation purposes.
The acquired knowledge is integrated as contextual guidance, enabling the LLM to produce translation candidates.
However, the generated knowledge may contain noise (see \cref{sec:noise-in-elicited-knowledge} for further analysis).
As a result, a filtering mechanism becomes necessary to select useful knowledge while filtering out unhelpful or noisy ones.

\subsection{Knowledge Mining}

Akin to the initial understanding and interpretation phase that human translators take~\cite{gile2009basic}, the knowledge mining step requires the LLM first to analyze the source text and elicit three aspects of knowledge generally beneficial to translation:

\textit{Keywords} are essential words or phrases that convey the core meaning of a text and act as focal points for understanding the main idea.
Accurate translation of keywords is crucial for conveying the intended meaning and ensuring faithfulness~\cite{baker2018other,koehn2009process}.
Besides, identifying and maintaining a list of keywords guarantees that specific terms are translated consistently across different parts of the text.

\textit{Topic} refers to the overall subject or theme being discussed.
A keen awareness of the topic helps translators sidestep potential issues arising from ambiguity, such as mistranslations or misinterpretations~\cite{bowker2002computer}.
It is important to highlight that topics are generally more specific than the broader domains that have been widely discussed within the machine translation community.
For example, while the news domain encompasses a wide range of subjects, subcategories like political news and entertainment news should adopt different registers and tones.

\textit{Demonstrations}, or example sentences, illustrate how comparable sentences can be translated accurately.
They assist the translators in identifying appropriate equivalents within the target language, enabling translators to produce natural and fluent translations to native speakers~\cite{hatim2004translation}.

As shown in Step 1 of \figurename~\ref{fig:method}, given the source sentence, we prompt the LLM to elicit keyword pairs, topics, and relevant demonstrations.\footnote{To ensure a uniform response format, we manually constructed 5-shot exemplars for each kind of knowledge.}

\subsection{Knowledge Integration}
Just as human translators weave their understanding of the source text into their translations~\cite{pym2014exploring}, knowledge integration embeds the acquired knowledge into the context (Step 2 in \figurename~\ref{fig:method}) and enables the LLM to utilize this information to generate multiple translation candidates.
We obtain four candidates, which the LLM generates without guidance from any \textit{external} knowledge.

\subsection{Knowledge Selection}
Knowledge selection resembles the final decision-making phase in human translation, where the best translation of the source text is chosen based on the context.
Although keywords, topics, and relevant demonstrations generally benefit translation, not all the LLM-generated knowledge is helpful.
For example, LLM may generate trivial or noisy content that might distract the translation process~\cite{shi2023large,agrawal-etal-2023-context}.
Our quantitative experiments in~\cref{sec:noise-in-elicited-knowledge} support this hypothesis.
Therefore, we employ a filtering mechanism to select the most useful knowledge and filter out the unhelpful or noisy ones.
Specifically, we adopt quality estimation (QE) to select the best candidate as the final output (Step 3 in \figurename~\ref{fig:method}).
The selection method is flexible, and both an externally trained QE model and the LLM itself served as QE are effective in our experiments.
\section{Experiments}

\subsection{Experimental Setup}
\begin{table*}[!htb] 
    \centering
    \resizebox{\linewidth}{!}{
    \begin{tabular}{l cc cc cc cc ccc}
        \toprule
       \bf Method                                                &\bf En-Zh &\bf Zh-En&\bf En-De&\bf De-En&\bf En-Ja&\bf Ja-En&\bf De-Fr&\bf Fr-De&\bf Cs-Uk&\bf Uk-Cs&\bf En-Hr\\
        \midrule
        \rowcolor{gray!25}
        \multicolumn{11}{c}{\textbf{\texttt{~~~~~~WMT22 Best | COMET}}} & \\
        \bf WMT22 Best                                           &    86.8  &    81.0 &   87.4  &    85.0 &    89.3 &    81.6 &    85.7 &    89.5 &    91.6 &    92.2 &    88.4\\
        \midrule
        \rowcolor{gray!25}
        \multicolumn{11}{c}{\textbf{\texttt{text-davinci-003 | COMET}}} & \\
          \bf Baseline                                           &    86.2  &    81.6 &   85.8  &    85.2 &    87.9 &    81.8 &    82.8 &    86.3 &    88.0 &    89.2 &    85.9\\
          \bf 5-Shot~(\citeauthor{hendy2023good})                &    87.0  &    81.1 &   86.5  &    85.2 &    88.2 &    82.0 &    83.6 &    86.6 &    ---- &    ---- &    ----\\
          \midrule
          \bf{Rerank}$_{\ \textsc{LLM-SCQ}}$                     &    86.4  &    81.7 &    86.0 &    85.2 &    88.0 &    82.0 &    83.0 &    86.4 &    88.3 &    89.4 &    86.3\\
          \bf{MAPS}$_{\ \textsc{LLM-SCQ}}$                       &    86.8  & \bf82.0 &    86.4 & \bf85.4 & \bf88.5 & \bf82.4 &    83.4 & \bf86.9 & \bf88.8 & \bf89.9 & \bf86.5\\
          \midrule
          \bf{Rerank}$_{\ \textsc{Comet-QE}}$                    &    86.9  & 82.1    &    86.4 &    85.5 &    88.8 &    82.3 &    83.4 &    86.8 &    89.4 &    90.1 &    87.1\\
          \bf{MAPS}$_{\ \textsc{Comet-QE}}$                      & \bf87.6  & \bf82.6 & \bf87.2 & \bf85.7 & \bf89.5 & \bf82.9 & \bf84.1 & \bf87.5 & \bf90.1 & \bf91.1 & \bf88.1\\
          \midrule
          $\overline{\Uparrow}$\ \bf{Rerank}$_{\ \textsc{Comet}}$&    87.5  &    82.6 &    86.9 &    85.8 &    89.3 &    82.3 &    83.4 &    86.8 &    89.9 &    90.7 &    87.7\\
          $\overline{\Uparrow}$\ \bf{MAPS}$_{\ \textsc{Comet}}$  & \bf88.5  & \bf83.8 & \bf88.0 & \bf86.7 & \bf90.3 & \bf82.9 & \bf84.1 & \bf87.5 & \bf90.9 & \bf92.0 & \bf89.0\\
          \midrule
        \rowcolor{gray!25}
        \multicolumn{11}{c}{\textbf{\texttt{~text-davinci-003 | BLEURT}}} & \\
          \bf Baseline                                           &    71.1  &    69.6 &    75.6 &    74.0 &    66.3 &    67.8 &    70.4 &    77.6 &    75.0 &    78.8 &    75.0\\
          \bf 5-Shot~(\citeauthor{hendy2023good})                &    72.2  &    69.2 &    76.3 &    74.5 &    67.1 &    68.0 &    70.9 &    78.0 &    ---- &    ---- &    ----\\
          \midrule
          \bf{Rerank}$_{\ \textsc{LLM-SCQ}}$                     &    71.4  &    69.8 &    75.9 &    74.1 &    66.6 &    68.1 &    70.6 &    77.7 &    75.3 &    79.0 &    75.4\\
          \bf{MAPS}$_{\ \textsc{LLM-SCQ}}$                       &    72.1  & \bf70.5 &    76.3 &    74.4 & \bf67.4 & \bf68.8 & \bf71.4 & \bf78.6 & \bf76.1 & \bf80.2 & \bf76.0\\
          \midrule
          \bf{Rerank}$_{\ \textsc{Comet-QE}}$                    &    71.7  &    70.1 &    76.1 &    74.3 &    67.3 &    68.3 &    71.2 &    78.1 &    76.4 &    79.7 &    75.9\\
          \bf{MAPS}$_{\ \textsc{Comet-QE}}$                      & \bf72.6  & \bf70.8 & \bf77.1 & \bf74.6 & \bf68.3 & \bf69.1 & \bf71.9 & \bf78.9 & \bf77.4 & \bf81.2 & \bf77.1\\
          \midrule
          $\overline{\Uparrow}$\ \bf{Rerank}$_{\ \textsc{Comet}}$&    72.4  &    70.6 &    76.5 &    74.6 &    68.0 &    68.8 &    71.8 &    78.6 &    76.8 &    80.2 &    76.4\\
          $\overline{\Uparrow}$\ \bf{MAPS}$_{\ \textsc{Comet}}$  & \bf74.0  & \bf72.1 & \bf77.8 & \bf75.7 & \bf69.4 & \bf70.9 & \bf73.6 & \bf80.2 & \bf78.3 & \bf82.1 & \bf77.9\\
          \midrule
       \rowcolor{gray!25}
       \multicolumn{11}{c}{\textbf{\texttt{~~~~~~~~~~Alpaca | COMET}}} & \\
          \bf Baseline                          &    58.9  &    73.1 &    75.5 &    81.9 &    56.6 &    71.8 &    71.7 &    75.4 &    74.1 &    71.1 &    65.9\\
          \midrule
          \bf{Rerank}$_{\ \textsc{Comet-QE}}$   &     66.2 &    74.9 &    78.5 &    82.6 &    64.7 &    73.7 &    74.5 &    78.2 &    78.1 &    76.3 &    70.5\\ 
          \bf{MAPS}$_{\ \textsc{Comet-QE}}$     &  \bf69.0 & \bf76.0 & \bf79.7 & \bf83.3 & \bf66.9 & \bf74.7 & \bf75.9 & \bf79.1 & \bf80.8 & \bf78.5 & \bf72.3\\
          \midrule
        \rowcolor{gray!25}
        \multicolumn{11}{c}{\textbf{\texttt{~~~~~~~~~~~Alpaca | BLEURT}}} & \\
          \bf Baseline                          &     42.3 &    58.0 &    62.2 &    69.8 &    31.4 &    55.4 &    52.2 &    63.4 &    52.4 &    54.3 &    53.2\\
          \midrule
          \bf{Rerank}$_{\ \textsc{Comet-QE}}$   &     47.5 &    59.5 &    64.7 &    70.4 &    36.2 &    56.7 &    55.0 &    66.0 &    55.2 &    59.0 &    56.0\\ 
          \bf{MAPS}$_{\ \textsc{Comet-QE}}$     &  \bf50.6 & \bf60.6 & \bf66.3 & \bf71.1 & \bf38.2 & \bf57.7 & \bf56.6 & \bf66.8 & \bf59.5 & \bf61.2 & \bf57.2\\
          \midrule
       \rowcolor{gray!25}
       \multicolumn{11}{c}{\textbf{\texttt{~~~~~~~~~~Vicuna | COMET}}} & \\
          \bf Baseline                          &     81.3 &    78.4 &    79.8 &    82.9 &    82.3 &    77.3 &    75.5 &    77.1 &    74.9 &    72.7 &    69.3\\
          \midrule
          \bf{Rerank}$_{\ \textsc{Comet-QE}}$   &     83.6 &    79.3 &    81.8 &    83.6 &    85.2 &    78.8 &    77.8 &    79.6 &    79.9 &    77.7 &    74.2\\
          \bf{MAPS}$_{\ \textsc{Comet-QE}}$     &  \bf84.5 & \bf80.2 & \bf82.7 & \bf84.1 & \bf86.5 & \bf79.7 & \bf79.2 & \bf81.1 & \bf81.8 & \bf80.1 & \bf76.0\\
          \midrule
        \rowcolor{gray!25}
        \multicolumn{11}{c}{\textbf{\texttt{~~~~~~~~~~~Vicuna | BLEURT}}} & \\
          \bf Baseline                          &     64.9 &    65.3 &    67.4 &    71.0 &    58.7 &    62.8 &    58.8 &    66.0 &    57.8 &    56.6 &    57.7\\
          \midrule
          \bf{Rerank}$_{\ \textsc{Comet-QE}}$   &     66.7 &    66.0 &    69.2 &    71.8 &    61.6 &    64.0 &    61.2 &    68.2 &    61.8 &    61.2 &    60.5\\
          \bf{MAPS}$_{\ \textsc{Comet-QE}}$     &  \bf67.8 & \bf66.9 & \bf70.0 & \bf72.4 & \bf63.0 & \bf64.8 & \bf62.5 & \bf69.3 & \bf64.0 & \bf64.3 & \bf63.4\\
       \bottomrule
    \end{tabular}}
    \caption{Translation performance on WMT22. \textbf{Bold} entries: denote statistically significant differences with $p < 0.05$ in the paired t-test compared to Baseline, 5-Shot and Rerank (with the same knowledge selection method). $\overline{\Uparrow}$: indicates the upper bound of selection, using \textsc{COMET}, a reference-based metric, as the selection method.}
    \label{tab:main-result}
\end{table*}

\paragraph{Models.}
We adopt three LLMs, encompassing both closed- and open-source models.

    \scalebox{0.8}{$\bullet$} text-davinci-003: A strong yet closed-source LLM developed by OpenAI, which employs advanced Reinforcement Learning with Human Feedback (RLHF) techniques~\cite{ouyang2022training}.
    We query it via the ofﬁcial API.

    \scalebox{0.8}{$\bullet$} Alpaca~\cite{alpaca}: An open-source and instruction-following LLM fine-tuned on LLaMA model~\cite{touvron2023llama1} with 52K Self-Instruct~\cite{selfinstruct} data.

    \scalebox{0.8}{$\bullet$} Vicuna~\cite{vicuna2023}: An open-source and instruction-following LLM fine-tuned on LLaMA-2 ~\cite{touvron2023llama2} with user-shared conversations collected from ShareGPT~\cite{sharegpt2023}.

For both Alpaca and Vicuna, we use the 7B version and perform inference on a single NVIDIA V100 32GB GPU.

\paragraph{Comparative Methods.}
For a rigorous comparison, we consider several variants, including single-candidate and multi-candidate methods.
Within single-candidate methods, we consider:
\begin{itemize}[leftmargin=10pt]
    \item[\labelitemi] \textbf{Baseline}: Standard zero-shot translation with temperature set to 0 (default value in this work).
    \item[\labelitemi] \textbf{5-Shot}~\cite{hendy2023good}: Five high-quality labeled examples from the training data are prepended to the test input, which performs best overall in \citet{hendy2023good}; meanwhile, increasing the number of examples will not result in meaningful improvement.
    This method requires meticulous construction of training data for each translation direction, including collecting, cleaning, and sorting by quality.
\end{itemize}
Within multi-candidate methods, we consider:
\begin{itemize}[leftmargin=10pt]
    \item[\labelitemi] \textbf{Rerank}: Using the same prompt as the Baseline, but with the temperature set to 0.3 (following~\citet{moslem2023adaptive}).
    We randomly sample three times and add Baseline to form four candidates.
    The best candidate is selected through QE.
    It can be considered as a pure reranking method without any guidance from extracted knowledge~\cite{fernandes-etal-2022-quality}.

    \item[\labelitemi] \textbf{MAPS}: Our proposed method described in Section~\ref{sec:MAPS}.
    Three translation candidates are generated with guidance from three aspects of knowledge.
    Combined with the Baseline, the best one is selected using QE.
\end{itemize}

\paragraph{Knowledge Selection Methods.}
\begin{itemize}[leftmargin=10pt]
    \item[\labelitemi] \textbf{LLM-SCQ}: Composing a single choice question (\textbf{SCQ}) that asks the LLM to choose the best candidate on its own.

    \item[\labelitemi] \textbf{\textsc{Comet-QE}}: A trained QE scorer that assigns a numerical score to each candidate.
    Selection is based on the highest score.

    \item[\labelitemi] \textbf{\textsc{Comet}} (oracle): A reference-based scorer that assigns a numerical score to each candidate.
    It can be considered as the oracle QE method, representing the upper bound of selection.
\end{itemize}

\paragraph{Test Data.} 
To avoid data leakage issues~\cite{bubeck2023sparks,garcia2023unreasonable,zhu2023multilingual}, we use the latest WMT22 test set, covering 11 translation directions at different resource levels (English $\Leftrightarrow$ Chinese, English $\Leftrightarrow$ German, English $\Leftrightarrow$ Japanese, German $\Leftrightarrow$ French, Ukrainian $\Leftrightarrow$ Czech and English $\Rightarrow$  Croatian).
WMT22 moves away from testing only on the news domain like in previous years and shifts to focus on the general scenario covering news, social, conversational, and e-commerce~\cite{kocmi-etal-2022-findings}.

\paragraph{Metrics.} 
We adopt COMET~\cite{rei-etal-2022-comet} and BLEURT~\cite{sellam-etal-2020-bleurt} as the main metrics.
These neural-based learned metrics show superiority over string-based metrics like BLEU~\cite{kocmi-etal-2021-ship,bawden2023investigating} and have been adopted broadly by LLM-based translation literature~\cite{moslem2023adaptive,gpt-mt-2023,garcia2023unreasonable,pilault2023interactive}.
We use wmt22-comet-da and BLEURT-20 checkpoints for these two metrics.

\subsection{Results}
For consistency, we are solely interested in comparing different methods under the same LLM.
As presented in \tablename~\ref{tab:main-result}, MAPS is broadly effective and exhibits a higher upper bound.
To be detailed, we have the following observations:

    \scalebox{0.8}{$\bullet$} \textbf{The effectiveness of MAPS has been validated across a wide range of settings.}
    Across 11 language pairs, 3 LLMs, and 2 metrics, MAPS consistently outperforms Rerank and Baseline.
    After employing MAPS$_{\ \textsc{Comet-QE}}$, text-davinci-003 surpasses the best submissions in WMT22 in 5 out of the 11 translation directions.
    This suggests that LLMs can enhance translation quality by emulating the human strategy of analyzing before translating.

    \scalebox{0.8}{$\bullet$} \textbf{MAPS outperforms Rerank consistently when the knowledge selection method is held constant.}
    This indicates that the improvements brought by MAPS stem from three types of translation-related knowledge: keywords, topics, and relevant demonstrations.
    We delve into the utilization of different types of knowledge and ablation study in \cref{sec: utilization-of-knowledge}.

    \scalebox{0.8}{$\bullet$} \textbf{Different knowledge selection methods can affect the final performance, and MAPS exhibits a higher upper bound for selection.}
    When using \textsc{LLM-SCQ}, the performance of MAPS is on par with 5-Shot ({MAPS}$_{\ {\textsc{LLM-SCQ}}}$ $\approx$ $\text{5-Shot}$); when using \textsc{Comet-QE}, MAPS consistently outperforms 5-Shot ({MAPS}$_{\ {\textsc{Comet-QE}}}$ $>$ $\text{5-Shot}$).
    More importantly, MAPS shows higher upper bounds for selection than Rerank ({MAPS}$_{\ {\textsc{Comet}}}$ $>$ {Rerank}$_{\ {\textsc{Comet}}}$), implying that superior knowledge selection methods like a better QE model~\cite{rei-etal-2022-cometkiwi}, AutoMQM~\cite{fernandes2023devil} or ranking strategy~\cite{fernandes-etal-2022-quality} can further improve MAPS.
\section{Analysis}
In this section, we conduct analyses to understand the MAPS framework.
If not otherwise specified, MAPS$_{\ \textsc{Comet-QE}}$, text-davinci-003, and WMT22 En-Zh are default tested method, model and language pair, respectively.

\subsection{Human Evaluation}
\label{sec:human-evaluation}
\begin{figure}[htpb]
    \centering
    \includegraphics[width=\linewidth]{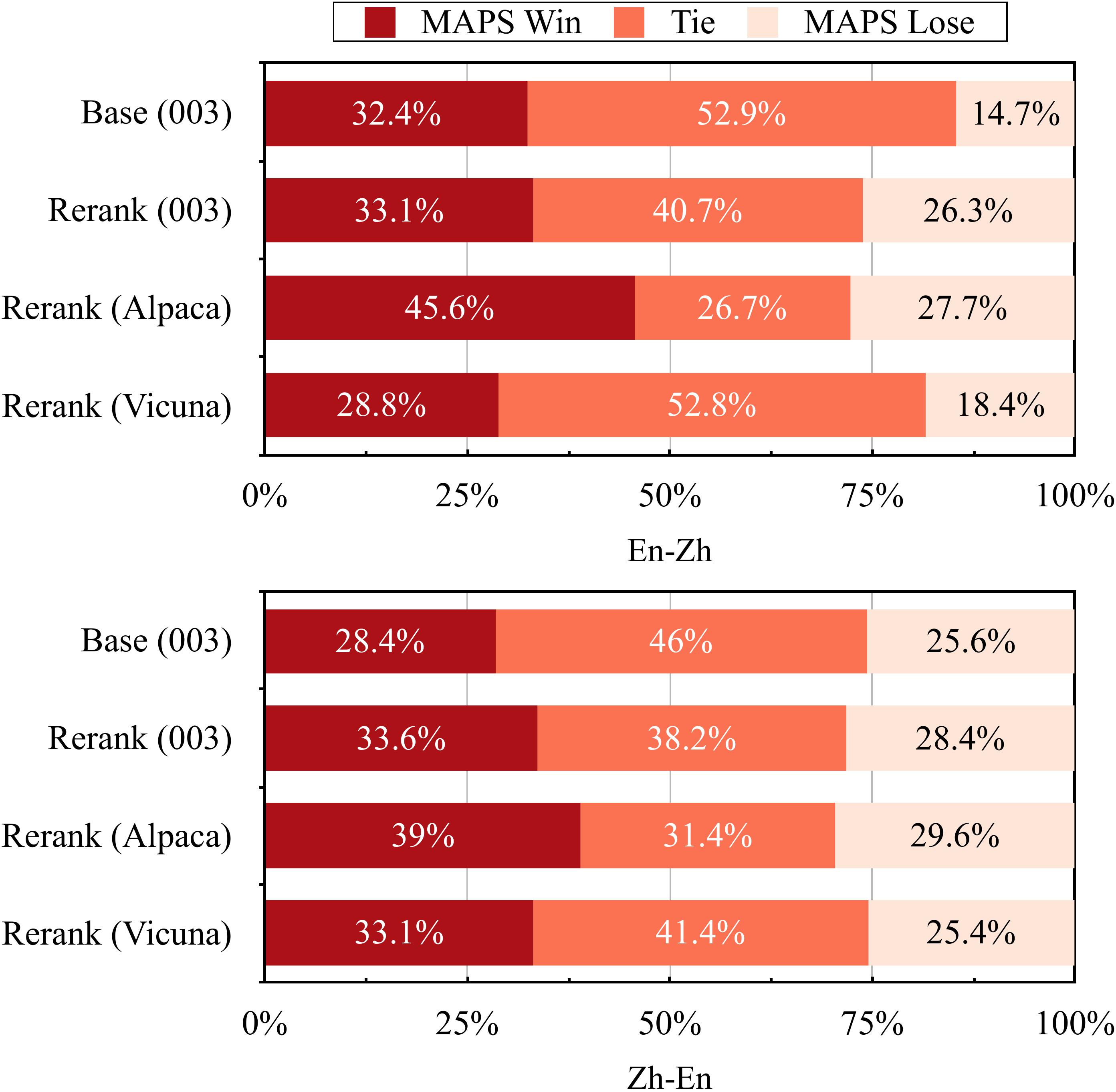}
    \caption{Human preference study, comparing MAPS with Base and Rerank. ``003'' denotes text-davinci-003.}
    \label{fig:human-preference}
\end{figure}

\paragraph{Preference Study.}
We perform human preference studies on En$\Leftrightarrow$Zh test sets.
For each test sample, our annotators (professional translators) were presented with a source sentence and two translations.
They were then tasked with selecting the superior translation or determining that neither translation was better than the other.
\figurename~\ref{fig:human-preference} shows the results of human preference studies, and MAPS is generally more preferred by humans.

\paragraph{MQM Evaluation.}
\begin{table}[b]
    \centering
    \begin{tabular}{lcc}
    \toprule
        \bf Method &\bf En-Zh&\bf Zh-En\\
    \midrule
        \bf Base&  1.94& 2.96\\ 
        \bf Rerank&  1.79& 2.84\\ 
        \bf MAPS&  \bf1.59& \bf2.60\\
    \bottomrule
    \end{tabular}
    \caption{Averaged MQM Score ($\downarrow$).}
    \label{tab:mqm-score}
\end{table}

\begin{figure}[t]
    \centering
    \includegraphics[width=\linewidth]{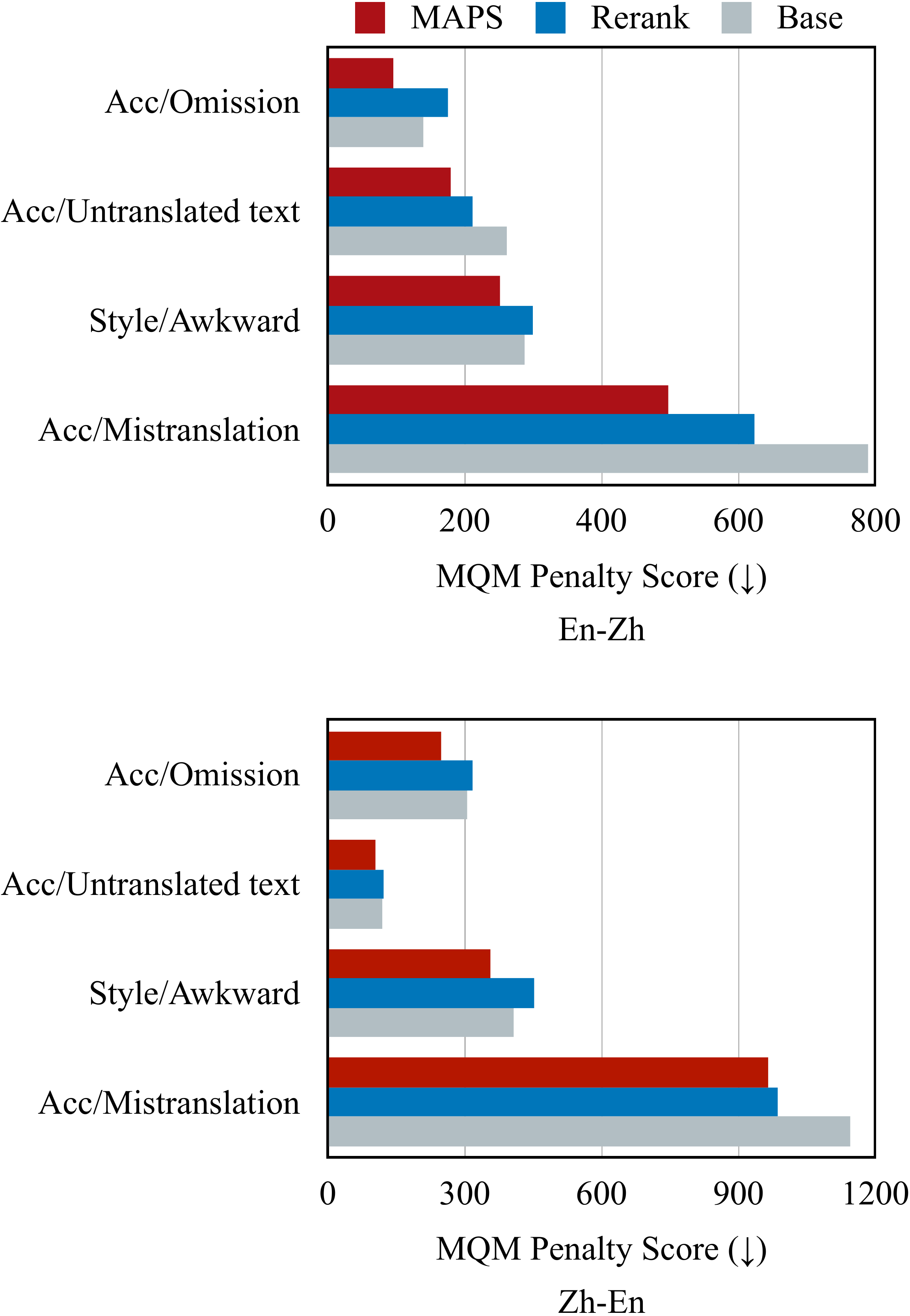}
    \caption{Selected MQM penalty scores (before average) under different error categories.}
    \label{fig:mqm-detail}
\end{figure}
To understand which aspects of translation that MAPS improves, we carried out MQM evaluations~\cite{burchardt-2013-multidimensional}.
MQM requires the annotators to identify the errors in translation and label the category and severity level for each error.
Based on the weights of the different error types, the MQM ends up with a penalty score.
We followed the assessment method in~\citet{freitag-etal-2021-experts}, including guidelines to annotators, error category, severity level and error weighting.
We employed professional translators who had MQM experience as the annotators.
We evaluated the first 1K samples on the Chinese$\Leftrightarrow$English test sets for cost reasons.
\tablename~\ref{tab:mqm-score} shows that MAPS outperforms Base and Rerank significantly.
In terms of error categories, the improvements brought about by MAPS are mainly in the reduction of mistranslation, awkward style, untranslated text, and omission errors, as presented in \figurename~\ref{fig:mqm-detail}.

\subsection{Utilization of Knowledge}
\label{sec: utilization-of-knowledge}
\begin{figure}[!htpb]
    \centering
    \includegraphics[width=\linewidth]{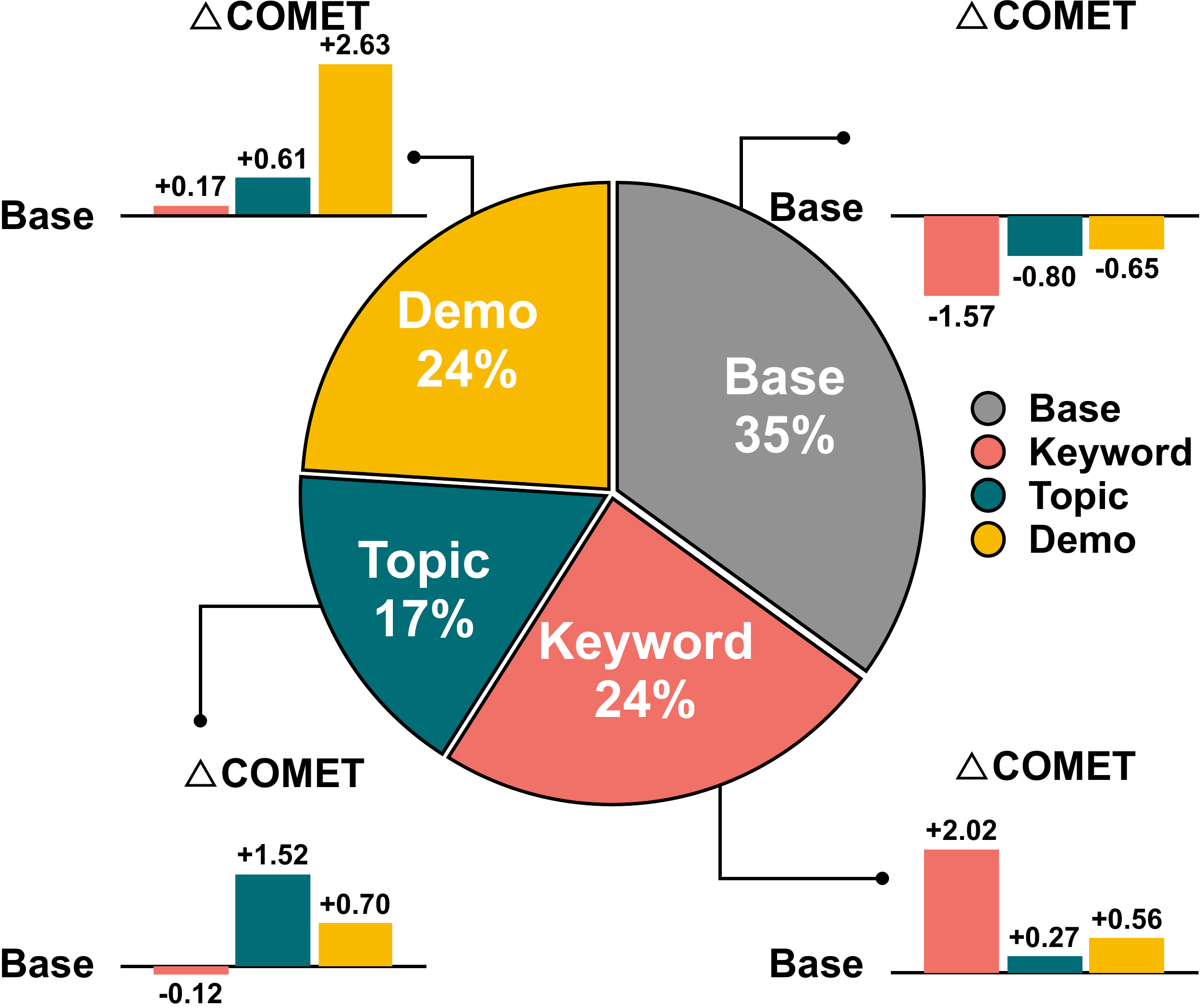}
    \caption{Utilization of keyword, topic and relevant demonstration in MAPS.}
    \label{fig:know-usage}
\end{figure}

\begin{table}[htpb]
    \small
    \centering
    \begin{tabular}{l l l}
    \toprule
        \bf Method                     & COMET &  BLEURT \\
    \midrule
        \textbf{Rerank} & 87.0      & 71.8 \\
    \hdashline
        \textbf{MAPS} & 87.6      & 72.6  \\
        \bf ~~- w/o Keyword            & 87.1$\textcolor{red}{_{\downarrow0.5}}$      & 72.1$\textcolor{red}{_{\downarrow0.5}}$ \\
        \bf ~~- w/o Topic              & 87.2$\textcolor{red}{_{\downarrow0.4}}$      & 72.4$\textcolor{red}{_{\downarrow0.2}}$ \\
        \bf ~~- w/o Demo               & 86.9$\textcolor{red}{_{\downarrow0.7}}$      & 72.0$\textcolor{red}{_{\downarrow0.6}}$ \\
    \bottomrule
    \end{tabular}
    \caption{Ablation study. We replace the knowledge-guided translation with random sampling translation in MAPS and report  average values of four experiments. ``$\textcolor{red}{\downarrow}$'': statistically significant difference with $p < 0.05$.}
    \label{tab:ablation}
\end{table}

Although \tablename~\ref{tab:main-result} reports the overall performance of MAPS, the utilization of the three aspects of knowledge remains unclear.
For instance, it is uncertain whether the majority of samples rely on relevant demonstrations rather than keywords and topics to guide the translation process.
To provide further insight, we illustrate the utilization of three types of knowledge in \figurename~\ref{fig:know-usage}.
We additionally present the performance differences among these three aspects of knowledge when applied to different subsets, relative to the baseline.
\figurename~\ref{fig:know-usage} reveals a relatively balanced utilization among them.
This implies that the three types of knowledge complement each other well within the MAPS framework.
The ablation study presented in \tablename~\ref{tab:ablation} further demonstrates the effectiveness of each type of knowledge.
Replacing any knowledge-guided translation with random sampling leads to performance degradation.

In \figurename~\ref{fig:know-usage}, we also note that the three types of knowledge cause different degrees of performance degradation when applied to the Base subset.
We conjecture that the knowledge elicited from the LLM is not always helpful and may even be noisy.
This finding motivates the knowledge selection step and is discussed in detail in \cref{sec:noise-in-elicited-knowledge}.

\subsection{Noise in Elicited Knowledge}
\label{sec:noise-in-elicited-knowledge}

The quality of extracted knowledge is essential for guiding translation.
We use keywords as an example to evaluate the quality of LLM-generated knowledge.
We design two metrics to characterize the quality of keyword pairs.
We denote $D=\{(s_i, t_i, h_i, K_i)\}$ as the set of test samples, where $s_i, t_i, h_i$ are source, target and hypothesis guided by the keyword pairs, respectively.
$K_i = \{(\mathrm{sw}_{ij}, \mathrm{tw}_{ij})\}$ denotes the LLM-generated keyword pairs for the $i$th sample, where
$(\mathrm{sw}_{ij}, \mathrm{tw}_{ij})$ is the $j$th keyword pair.
The precisions of LLM-generated keyword pairs concerning the source or target are defined as:\footnote{For simplicity's sake, we use subset notation to represent substring relationship.}
\begin{align}
    P_{\mathrm{src}} &= \frac{\sum_i\sum_j \mathbf{1}(\mathrm{sw}_{ij} \subseteq	 s_i)}{\sum_{i}|K_i|}, \\
    P_{\mathrm{tgt}} &= \frac{\sum_i\sum_j \mathbf{1}(\mathrm{tw}_{ij} \subseteq	 t_i)}{\sum_{i}|K_i|},  
\end{align}
where $\mathbf{1}(\cdot)$ denotes the indicator function.
$P_{\mathrm{src}}$ and $P_{\mathrm{tgt}}$ reflect the proportion of LLM-generated keyword pairs that do exist in the source and target, respectively.
Similarly, to evaluate how well the model follows the given keyword pairs, we define the recall of keywords in the LLM hypothesis:
\begin{align}
    R &= \frac{\sum_i\sum_j \mathbf{1}(\mathrm{tw}_{ij} \subseteq	 h_i)}{\sum_{i}|K_i|}.
\end{align}

The statistical results in \tablename~\ref{tab:kw-quality} show that: (1) although most LLM-generated keywords appear in the source sentences, only about half of them appear in the target sentences (55.8\% for En-Zh; 41.8\% for Zh-En).
(2) the LLM strictly follows the given keyword pairs when performing translations (97.1\% for En-Zh; 89.5\% for Zh-En).

Combining the above two observations, we can conclude that the LLM-generated knowledge contains a certain degree of noise (at least content that is not consistent with the reference), which can easily mislead the translation process.
This explains why incorporating that knowledge in the ``Base'' part of \figurename~\ref{fig:know-usage} brings negative effects.
Hence, knowledge selection is a crucial step in the MAPS framework to reduce the impact of noise.
\begin{table}[htpb]
    \centering
    \begin{tabular}{ccc ccc}
    \toprule
    \multicolumn{3}{c}{\bf En-Zh} & \multicolumn{3}{c}{\bf Zh-En} \\
    \cmidrule(lr){1-3} \cmidrule(lr){4-6} 
    $P_{src}$ & $P_{tgt}$ & $R$ & $P_{src}$ & $P_{tgt}$ & $R$ \\
    \midrule
     98.8 & 55.8 & 97.1 & 99.2 & 41.8 & 89.5  \\
    \bottomrule 
    \end{tabular}
    \caption{Quality of LLM-generated keyword pairs.}
    \label{tab:kw-quality}
\end{table}

\begin{table}[htpb]
    \centering
    \begin{tabular}{l c c c}
    \toprule
        \bf{Method} & COMET & BLEURT & Accuracy \\
    \midrule
       \bf{Rerank}  & 81.5  & 70.2 & 61.5 \\
       \bf{MAPS}    & \bf82.2  & \bf70.6 & \bf65.5 \\ 
    \bottomrule
    \end{tabular}
    \caption{Results on lexical ambiguity test set.}
    \label{tab:ambiguity}
\end{table}

\subsection{MAPS Helps Ambiguity Resolution}
\label{sec:maps-helps-ambiguity-resolution}
Ambiguity resolution has long been one of the most challenging problems in machine translation.
To evaluate the ambiguity resolution capability of machine translator, \citet{he-etal-2020-box} provides a lexical ambiguity test set for Chinese$\rightarrow$English.
The hard part of this test set involves Chinese sentences which are difficult to translate correctly unless the translator resolves their ambiguities.
Our test results in \tablename~\ref{tab:ambiguity} show the superiority of MAPS in ambiguity resolution, where the ``accuracy'' indicates the percentage of successfully disambiguated sentences (evaluated by human).

\subsection{MAPS Reduces LLM's Hallucinations}
\label{sec:maps-reduces-llms-hallucinations}
Hallucination issue in natural language generation (NLG) refers to the phenomenon \emph{where the content generated by the model is nonsensical or unfaithful to the provided source content}~\cite{10.1145/3571730,filippova-2020-controlled,maynez-etal-2020-faithfulness,parikh-etal-2020-totto,zhou-etal-2021-detecting,he-etal-2022-bridging}.
This has been one of the key challenges in LLMs~\cite{zhang2023multicot}.
In this section, we analyze the phenomenon of hallucination through automatic and human evaluation.

\begin{table}[htpb]
    \centering
    \begin{tabular}{l  l  l}
            \toprule
           \bf Method                                & $\Delta$\% hallucinations \\
            \midrule
              \bf Baseline                           & --\\
              \midrule
              \bf{Rerank}$_{\ \textsc{Comet-QE}}$    & -3\% \\
              \bf{MAPS}$_{\ \textsc{Comet-QE}}$    & -8\% \\
              \midrule
              $\overline{\Uparrow}$\ \bf{Rerank}$_{\ \textsc{Comet}}$       & -6\% \\
              $\overline{\Uparrow}$\ \bf{MAPS}$_{\ \textsc{Comet}}$       & -12\% \\
           \bottomrule
    \end{tabular}
    \caption{$\Delta$\% of token-level hallucinations. $\overline{\Uparrow}$: indicates the upper bound of selection, using \textsc{COMET}, a reference-based metric, as the selection method.}
    \label{tab:token-level-hallu}
\end{table}

In automatic evaluation, we use the hallucination detector provided by~\citet{zhou-etal-2021-detecting} to identify token-level hallucination in Alpaca's translation on Chinese$\rightarrow$English test set.
The detector assigns a binary label to each generated token.
In \tablename~\ref{tab:token-level-hallu}, MAPS outperforms Rerank and demonstrates a higher upper bound.

\begin{figure}
    \centering
    \includegraphics[width=0.9\linewidth]{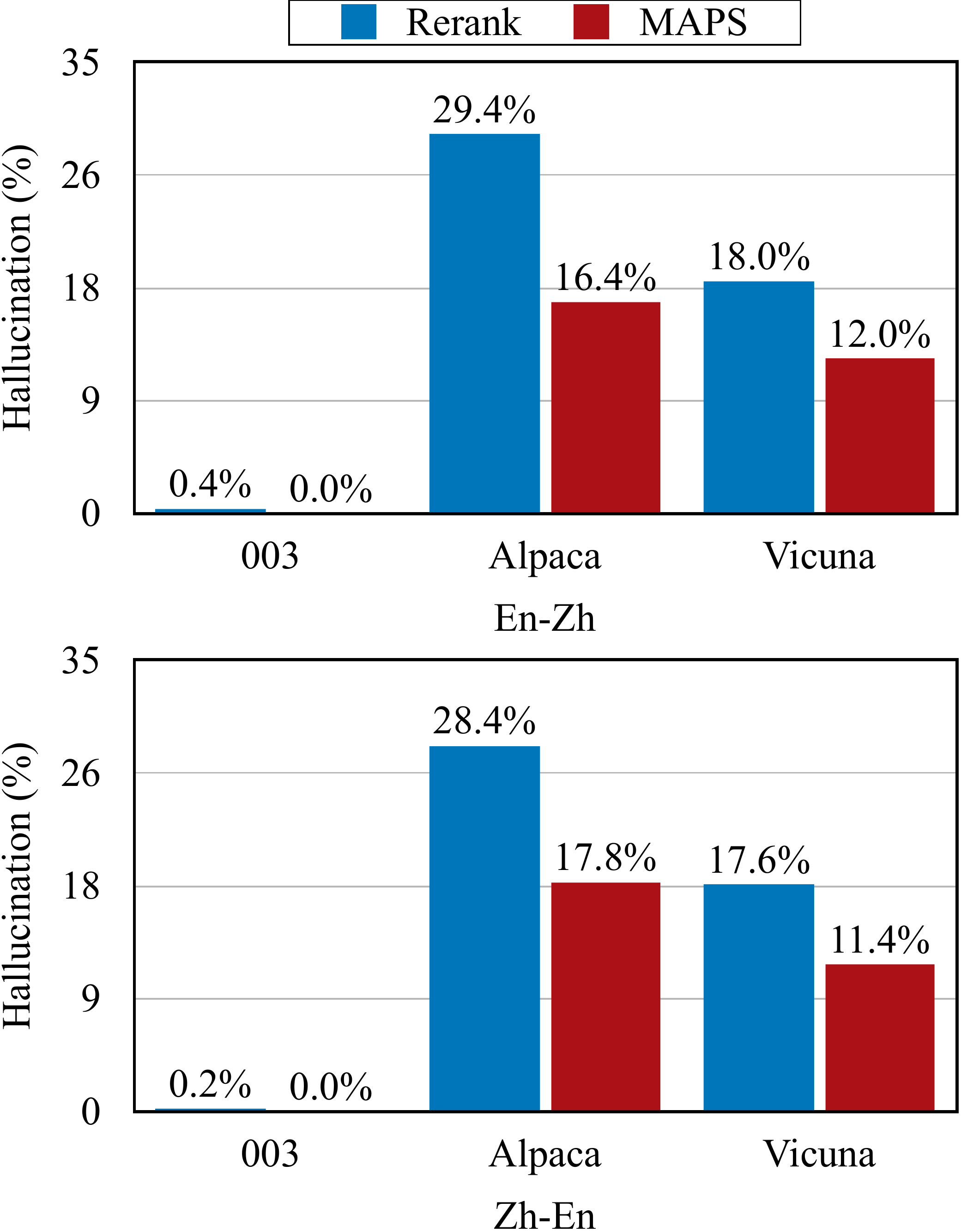}
    \caption{Ratio of hallucinations. Human annotators were tasked with labeling whether the generated translation is a hallucination error. ``003'' denotes text-davinci-003 and it has almost no hallucination errors.}
    \label{fig:hallu-human}
\end{figure}

In human evaluation, we employed professional human translators to label the hallucination errors in both MAPS and Rerank.
We sampled 500 sentences from each of the English$\Leftrightarrow$Chinese test sets and evaluated text-davinci-003, Alpaca, and Vicuna.
The human annotators were required to decide whether the translation belongs to the hallucination error following the definition from \citet{guerreiro-etal-2023-looking}.
The results from \figurename~\ref{fig:hallu-human} show that MAPS outperforms Rerank by a notable margin in resolving hallucination.

We conjecture that one of the key differences between MAPS and Rerank is that MAPS is enabled to correct the probability distribution of the next token prediction, while Rerank is not.
If a hallucinatory token occupies a high probability mass~\cite{Wang2022UnderstandingAI}, it is difficult for Rerank to avoid selecting this token by diverse sampling.
In contrast, MAPS, providing additional translation-related knowledge in the prompt, enables the model to redistribute the probability of the next token, thus offering more possibilities to avoid choosing the hallucinatory token.

\subsection{Three-in-One Prompting}
So far, we have discussed the case where the LLM uses the three types of knowledge separately.
An immediate question is how the LLM would perform if the three types of knowledge were integrated into one prompt.
We call this method three-in-one prompting and present results in \tablename~\ref{tab:three-in-one-prompting}.

\begin{table}[htpb]
    \centering
    \resizebox{\linewidth}{!}{
    \begin{tabular}{l ll ll}
    \toprule
    \bf Method & \bf En-Zh &\bf Zh-En & \bf En-De &\bf De-En\\
    \midrule
    \rowcolor{gray!25}
    \multicolumn{5}{c}{\textbf{\texttt{text-davinci-003 | COMET~~}}} \\
    \bf Baseline      &     86.2         & 81.6         & 85.8         & 85.2        \\
    \bf Three-in-One  &     86.7\ii{0.5} & 81.7\ii{0.1} & 86.1\ii{0.3} & 85.1\dd{0.1}\\
    \midrule
    \bf{MAPS}$_{\ \textsc{Comet-QE}}$    & 87.6         & 82.6         &     87.2         & 85.7        \\
    \bf{MAPS}$^+_{\ \textsc{Comet-QE}}$  & 87.6\nn{0.0} & 82.6\nn{0.0} &     87.3\ii{0.1} & 85.7\nn{0.0}\\
    \midrule
    \rowcolor{gray!25}
    \multicolumn{5}{c}{\textbf{\texttt{text-davinci-003 | BLEURT}}} \\
    \bf Baseline      &     71.1         &     69.6         &     75.6         & 74.0\\
    \bf Three-in-One  &     71.8\ii{0.7} &     70.1\ii{0.5} &     76.1\ii{0.5} & 74.0\nn{0.0}\\
    \midrule
    \bf{MAPS}$_{\ \textsc{Comet-QE}}$    & 72.6         & 70.8         &     77.1         & 74.6\\
    \bf{MAPS}$^+_{\ \textsc{Comet-QE}}$  & 72.6\nn{0.0} & 70.9\ii{0.1} &     77.2\ii{0.1} & 74.6\nn{0.0}\\
    \midrule
    \midrule
             & \bf En-Ja &\bf Ja-En & \bf De-Fr &\bf Fr-De\\
    \midrule
    \rowcolor{gray!25}
    \multicolumn{5}{c}{\textbf{\texttt{text-davinci-003 | COMET~~}}} \\
    \bf Baseline     & 87.9         & 81.8         & 82.8         & 86.3\\
    \bf Three-in-One & 88.4\ii{0.5} & 81.9\ii{0.1} & 83.1\ii{0.3} & 86.5\ii{0.2}\\
    \midrule
    \bf{MAPS}$_{\ \textsc{Comet-QE}}$    & 89.5         & 82.9         & 84.1         & 87.5\\
    \bf{MAPS}$^+_{\ \textsc{Comet-QE}}$  & 89.6\ii{0.1} & 83.0\ii{0.1} & 84.2\ii{0.1} & 87.6\ii{0.1}\\
    \midrule
    \rowcolor{gray!25}
    \multicolumn{5}{c}{\textbf{\texttt{text-davinci-003 | BLEURT}}} \\
    \bf Baseline     & 66.3         & 67.8         & 70.4         & 77.6\\
    \bf Three-in-One & 67.3\ii{1.0} & 68.3\ii{0.5} & 70.6\ii{0.2} & 78.0\ii{0.4}\\
    \midrule
    \bf{MAPS}$_{\ \textsc{Comet-QE}}$    & 68.3         & 69.1         & 71.9         & 78.9\\
    \bf{MAPS}$^+_{\ \textsc{Comet-QE}}$  & 68.5\ii{0.2} & 69.3\ii{0.2} & 71.9\nn{0.0} & 79.0\ii{0.1}\\
    \bottomrule
    \end{tabular}}
    \caption{Three-in-one prompting. Three-in-One: three types of knowledge are integrated into one prompt. {MAPS}$^+_{\ \textsc{Comet-QE}}$: adding candidate produced by Three-in-One into {MAPS}$_{\ \textsc{Comet-QE}}$. The subscripts indicate relative improvements from three-in-one prompting.}
    \label{tab:three-in-one-prompting}
\end{table}

Within single-candidate methods (Baseline v.s. Three-in-One), three-in-one prompting brings positive results overall, which means that the LLM can use three types of knowledge simultaneously.
However, the degree of improvement varies significantly under different language pairs, with notable absence of effect in De-En translation.
Regarding multi-candidate methods (MAPS$_{\ \textsc{Comet-QE}}$ v.s. MAPS$^+_{\ \textsc{Comet-QE}}$), incorporating three-in-one prompting into MAPS yields only marginal improvements ($\leq$0.2).
Considering that the candidate set generated by the three-in-one prompting overlaps significantly with the candidate sets generated individually by the three types of knowledge, this result is as expected.
\section{Related Work}

\subsection{LLMs for Translation}
Research evaluating the translation capabilities of LLMs falls into two main lines.
The first line involves issues specific to LLMs, including the impact of demonstration selection in ICL~\cite{vilar2022prompting,zhang2023prompting,garcia2023unreasonable} and prompt templates~\cite{zhang2023prompting,jiao2023ischatgpt} on translation performance.
The second line focuses on comprehensive evaluations of LLMs under various translation scenarios, covering multilingual~\cite{jiao2023ischatgpt,zhu2023multilingual,gpt-mt-2023}, document-level~\cite{gpt-mt-2023,wang2023document,karpinska2023large}, low-resource translation~\cite{jiao2023ischatgpt,garcia2023unreasonable,zhu2023multilingual,bawden2023investigating}, robustness~\cite{jiao2023ischatgpt}, hallucination~\cite{guerreiro2023hallucinations} and domain adaptation~\cite{gpt-mt-2023,wang2023guofeng}.
Our work evaluates the translation capabilities of LLMs across eleven translation directions, varying from same-family (En$\Leftrightarrow$De), distant (En$\Leftrightarrow$Ja, En$\Leftrightarrow$Zh) and non-English-centric (De$\Leftrightarrow$Fr) and low-resource (Cs$\Leftrightarrow$Uk, En$\Rightarrow$Hr) language pairs.
\citet{zhu2023multilingual} emphasizes the risk of data leakage.
Therefore, we adopt the latest WMT22 test sets.
Our work also quantitatively evaluates ambiguity resolution and token-/sentence-hallucination in LLM-based translation.

\citet{jiao2023parrot} incorporates human evaluation into instruction data for training, resulting in translations that are preferred by humans during interactive chat sessions. 
In contrast, our work takes a different approach by mimicking the human translation process and achieves higher-quality translations without training.

\citet{agrawal-etal-2023-context} proposes an algorithm based on n-gram recall for demonstration selection.
Given the ground-truth context, \citet{pilault2023interactive} introduces an interactive-chain prompting method for ambiguity resolution.
\citet{moslem2023adaptive} suggests prompting the LLMs with terminology hints extracted from the selected demonstrations or a compiled glossary for domain-specific translation such as COVID-19.
Concurrently, \citet{ghazvininejad2023dictionary} and \citet{lu2023chain} use external dictionaries to augment prompts for low-resource and domain-specific translation.
While our work can be viewed as a form of ``prompting strategy'', it differs from this line of research in that it does not rely on any external ``datastore'', such as sample pools, dictionaries or ground-truth context, which should be curated carefully for specified language pairs or domains.
In contrast, we consider the LLM itself as a ``datastore'' containing broad knowledge that can assist its translation process.

\subsection{Chain-of-Thought Prompting}
\citet{wei2022chain} explores how chain-of-thought (CoT) prompting improves the ability of LLMs to perform complex reasoning such as arithmetic reasoning, commonsense reasoning, and symbolic reasoning. 
By guiding LLMs through generating intermediate reasoning chains prior to reaching a final solution, CoT prompting has propelled the multi-step reasoning abilities of LLMs to an extraordinary level, as substantiated by previous research \citep{wei2022emergent,wang2023self}.
CoT prompting manifests through two distinct paradigms, namely zero-shot CoT \citep{zero-shot,yang2023large} and few-shot CoT \citep{wei2023chainofthought,zhang2023automatic}.
Zero-shot CoT simply appends a trigger prompt such as \emph{Let's think step by step} after the test question, with the motivation to harness the step-by-step reasoning capacities of LLMs in a zero-shot manner. Few-shot CoT operates by utilizing a few input-output demonstrations, each of which comprises a question, a reasoning chain, and the corresponding answer. These demonstrations are seamlessly integrated before the test question, resulting in a prompted input that is subsequently processed by an LLM to deduce the answer. 

So far, most CoT prompting studies focus on complex reasoning problems. Although there are a few preliminary attempts to extend CoT prompting techniques to machine translation tasks, 
\citet{Peng2023ChatGPT4MT} finds that straightforwardly applying CoT to translation tasks resulted in word-by-word translations, which is less than satisfactory. Following this line, our work can also be viewed as a form of CoT prompting for translation as it dissects the translation process into distinct steps, which is the first successful attempt of CoT in translation tasks to the best of our knowledge. Notably, our work has successfully achieved improved translation performance by inducing three aspects of translation-related knowledge including keywords, topics, and relevant demonstrations to guide the final translation process.

\subsection{Self-Prompting}
Self-prompting is a line of research that utilizes the LLMs to prompt themselves and extract relevant knowledge to aid downstream tasks\citep{li2022self,wang-etal-2023-element}. Diverging from CoT prompting which focuses on providing intermediate reasoning steps on the output side, self-prompting techniques dissect the input problem into specific sub-problems on the input side and extract the salient knowledge for the sub-problems one by one. This extracted knowledge is then utilized to deduce the ultimate solution. 

Several studies exemplify the diversity of self-prompting applications. Specifically, 
\citet{kim2022self} and \citet{li2022self} use the LLMs to generate in-context exemplars for text classification and open-domain question answering, respectively.
\citet{yu2023generate} generates diverse documents from the LLMs to improve knowledge-intensive tasks. \citet{wang-etal-2023-element} compels LLMs to first extract the core elements for news texts, such as entity, date, event, and result. Then, the extracted elements are used to generate summaries.
Further innovations emerge in multimedia contexts. \citet{zhu2023chatgpt} and \citet{chen2023video} empower LLMs to pose inquiries regarding provided images and videos to enrich the caption.
Remarkably, MAPS extends the domain of self-prompting into machine translation for the first time.
\section{Conclusion}
This work introduces MAPS, a method that enables LLMs to mimic human translation strategy for achieving high-quality translation.
MAPS allows LLMs to take preparatory steps before translation.
Specifically, LLMs analyze the given source text and generate three aspects of translation-related knowledge: keywords, topics, and relevant demonstrations.
Using a filtering mechanism based on quality estimation, the selected knowledge guides the LLMs' translation process.
In experiments with text-davinci-003, Alpaca and Vicuna, MAPS yields significant and consistent improvements across eleven translation directions from WMT22 and exhibits a higher upper bound of candidate selection.
Human evaluations show that MAPS provides more favorable translations by reducing mistranslation, awkward style, untranslated text, and omission errors.
Further analyses show that MAPS effectively resolves ambiguities and hallucinations in translation.
Future work includes designing more aspects of translation-related knowledge and better filtering mechanisms to improve the translation capabilities of LLMs further.
Another interesting direction is to explore the human-like translation strategy in training LLMs (e.g., instruction tuning).
\section{Discussion}
\label{sec:discussion}
\subsection{Inference Time}
\begin{figure}[htpb]
    \centering
    \includegraphics[width=\linewidth]{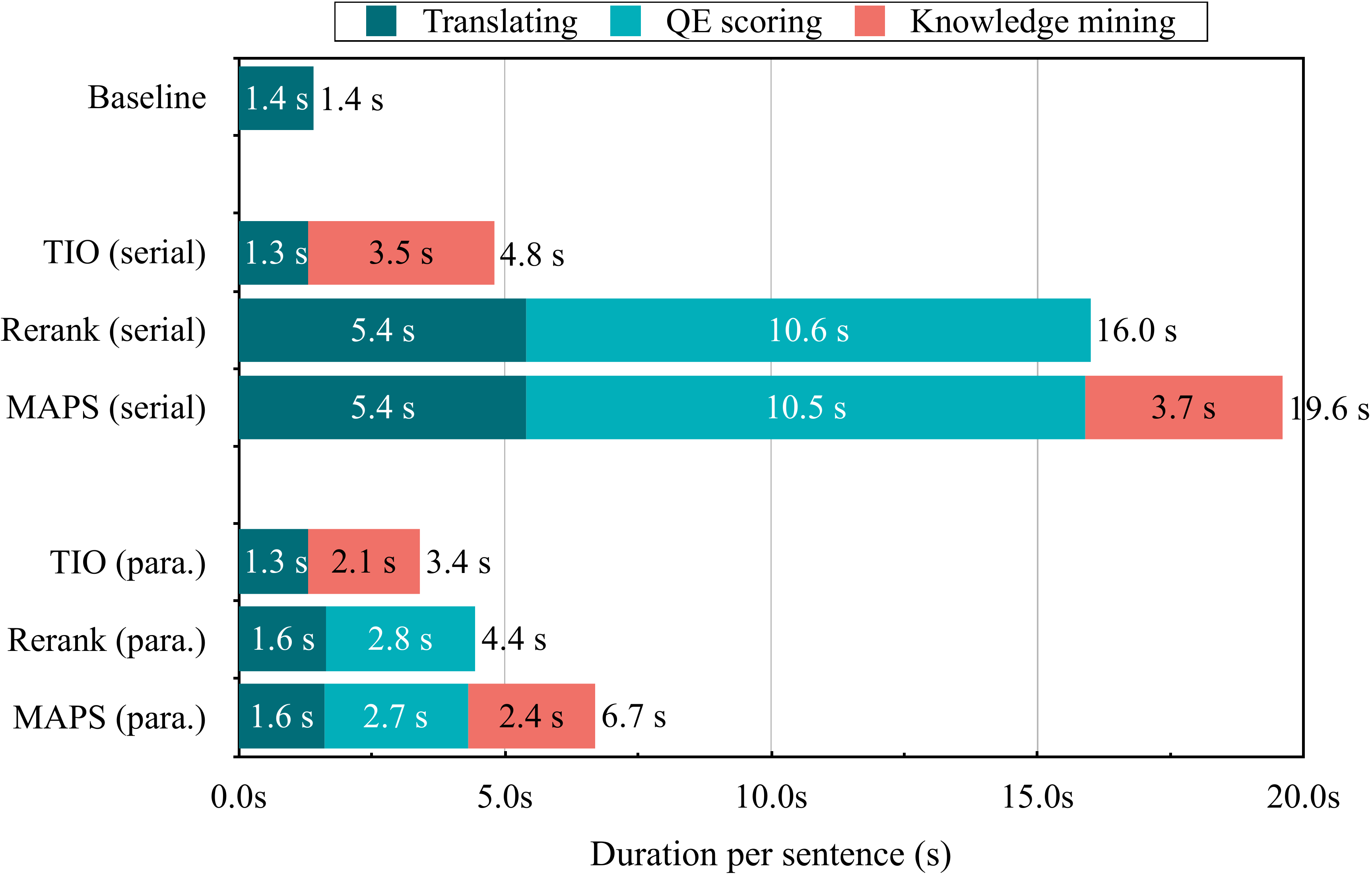}
    \caption{Durations of different processing stages for Baseline, There-in-One prompting (TIO), Rerank, and MAPS. The results represent the average of five independent trials. Experiments were conducted using a T4-8C GPU, with each trial consisting of 50 sentences. ``Serial'' and ``Para.'' denote serial and parallel processing of multiple types of knowledge and candidates, respectively.}
    \label{fig:time}
\end{figure}
Since MAPS consists of three sequential stages, the main limitation of MAPS lies in inference time.
As shown in \figurename~\ref{fig:time}, when processing serially, the inference times of Three-in-One, Rerank, and MAPS are 3$\times$, 11$\times$, and 14$\times$ the Baseline, respectively.

Given that all three methods involve processing multiple types of knowledge or candidates without any dependencies between them, a practical approach for acceleration is to parallel processing, which drastically reduces the running times (\DD{29\%} for Three-in-One; \DD{73\%} for Rerank; \DD{66\%} for MAPS) to an acceptable level.

The additional overhead from MAPS is mainly in the knowledge mining phase, where the LLM generates three types of knowledge separately.
One possible acceleration is to have the LLM generate three types of knowledge in a single call.
By controlling the format of the output, e.g. JSON, we can extract each type of knowledge.
However, the LLM is not guaranteed to output valid JSON content, which may lead to degradation of the final translation performance (see \tablename~\ref{tab:json}).

In addition, the running time of the QE scoring can be reduced by techniques such as model quantization or compression.
\begin{table}[htpb]
    \centering
    \resizebox{\linewidth}{!}{
    \begin{tabular}{l lll lll}
    \toprule
    \multirow{2}{*}{\bf Method}     & \multicolumn{3}{c}{\bf En-Zh} & \multicolumn{3}{c}{\bf Zh-En} \\
    \cmidrule(lr){2-4} \cmidrule(lr){5-7}
               & CT & BT & JSON E. & CT & BT & JSON E. \\
    \midrule
    \bf{MAPS}  & 87.6 & 72.6 & --- & 82.6 & 70.8 & --- \\
    \bf{MAPS}$^{\textsc{JSON}}$ & 87.7 & 72.6 & 0.1\% & 82.1\dd{} & 70.3\dd{} & 2.0\% \\
    \bottomrule
    \end{tabular}}
    \caption{{MAPS}$^{\textsc{JSON}}$: generating three types of knowledge in one JSON object. CT: COMET; BT: BLEURT. JSON E.: percentage of model output that is not in valid JSON format. Times taken for knowledge mining using MAPS and {MAPS}$^{\textsc{JSON}}$ are 3.7 and 2.3 seconds per sentence, respectively.}
    \label{tab:json}
\end{table}

\subsection{Is MAPS Overfitting Evaluation Metrics?}
In this work, we rely on COMET and BLEURT for automatic evaluation for their strong alignment with human evaluation, as highlighted by~\citet{freitag-etal-2022-results}.
We also use COMET-QE as one of the knowledge selection methods, whose training data has overlap with evaluation metrics.
This leads to a pertinent question: is MAPS merely overfitting to COMET and BLEURT?

To ensure reliable evaluations, we integrated human assessments into all our experiments, including: MQM evaluation (\cref{sec:human-evaluation}), human preference studies (\cref{sec:human-evaluation}), ambiguity resolution (\cref{sec:maps-helps-ambiguity-resolution}), and analysis of hallucination (\cref{sec:maps-reduces-llms-hallucinations}).
These evaluations substantiate MAPS's effectiveness from the viewpoint of human translators.

Furthermore, we demonstrate that MAPS remains effective even in the absence of COMET-QE.
As shown in Table~\ref{tab:main-result}, by formulating single-choice questions, the LLM itself can select the best translation candidates (Rerank$_{\ \textsc{LLM-SCQ}}$ and MAPS$_{\ \textsc{LLM-SCQ}}$).

\begin{table}[htpb]
    \centering
    \resizebox{\linewidth}{!}{
    \begin{tabular}{c|p{3.7cm}|c c c}
    \hline
    \bf Error type & \bf Translation & \makecell{\bf COMET-QE \\ (ref-free)} & \makecell{\bf COMET \\ (ref-based)} & \makecell{\bf BLEURT \\ (ref-based)} \\
    \hline
    \makecell{None \\ (the reference)} & The rule of drinking Red Label Whisky:     &  2.8 & \bf 96.2 & \bf 101.8\\
    \hline
    \makecell{Hallucination \\ (irrelevant content)} & The rule of drinking Red Label Whisky: \hl{1. Always drink responsibly.2. Never drink alone.3. Avoid drinking on an empty stomach.4. Set limits and stick to them.5. Drink in moderation.} & \bf 5.6 & 64.1 & 70.2\\
    \hline
    \makecell{Off-target \\ (wrong target language)} &  \hl{So trinkt man Red-Label-Whisky:} & \bf 4.6 & 72.6 & 53.2\\
    \hline
    \end{tabular}
    }
    \caption{Case study of Chinese-to-English translation for ``\zh{红牌威士忌喝法：}''. COMET-QE assigns higher scores to translations with hallucination and off-target errors compared to an error-free translation. Conversely, COMET and BLEURT award lower scores to these erroneous translations. Error spans are \hl{highlighted}.}
    \label{tab:error-cases}
    \end{table}

From a data perspective, all three models above were trained using datasets from WMT.
However, they use the data in different ways.
COMET-QE is reference-free and does not utilize reference data during training or inference.
On the contrary, COMET and BLEURT are reference-based, with both training and inference processes relying on reference data.
This difference allows COMET and BLEURT to penalize translation errors against a reference, a function that COMET-QE lacks due to its reference-free design (see \tablename~\ref{tab:error-cases}).

Overall, MAPS is widely effective by employing human strategy for translation.

\section*{Acknowledgements}
Zhiwei and Rui are with MT-Lab, Department of Computer Science and Engineering, School of Electronic Information and Electrical Engineering, and also with the MoE Key Lab of Artificial Intelligence, AI Institute, Shanghai Jiao Tong University, Shanghai 200204, China.  Rui and Zhiwei are supported by the Tencent Open Fund (RBFR2023012), the National Natural Science Foundation of China (62176153), and the Shanghai Municipal Science and Technology Major Project (2021SHZDZX0102).

We are grateful to the action editor and reviewers, whose insightful suggestions and exceptionally prompt feedback significantly enhanced the quality of our manuscript.

\appendix
\section{Knowledge-specific Prompting for Rerank}
\begin{table}[!htb] 
    \centering
    \resizebox{\linewidth}{!}{\begin{tabular}{l cc cc}
       \toprule
       \multirow{2}{*}{\bf Method}                               &\multicolumn{2}{c}{\bf En-Zh} &\multicolumn{2}{c}{\bf Zh-En}\\
       \cmidrule(lr){2-3} \cmidrule(lr){4-5}
       & COMET & BLEURT & COMET & BLEURT \\
        \midrule
       \rowcolor{gray!25}
       \multicolumn{5}{c}{\textbf{\texttt{text-davinci-003}}}\\

          \bf Baseline                                   &    86.2  &    71.1  &    81.6  &    69.6\\
          \bf{Rerank}$_{\ \textsc{}}$                    &    86.9  &    71.7  &    82.1  &    70.1\\
        \midrule
          \bf{Rerank}$_{\ \textsc{Keyword}}$             &    86.7  &    71.9  &    81.8  &    70.2\\
          \bf{Rerank}$_{\ \textsc{Topic}}$               &    87.1  &    71.8  & \bf82.2  & \bf70.5\\
          \bf{Rerank}$_{\ \textsc{Demo}}$                & \bf87.4  & \bf72.7  &    82.1  &    70.4\\
       \bottomrule
    \end{tabular}}
    \caption{Knowledge-specific prompting for Rerank. Four hypotheses are sampled from each knowledge-specific prompt and reranked. The subscript indicates the type of knowledge.}
    \label{tab:know-spec-rerank}
\end{table}

\bibliography{anthology,custom}
\bibliographystyle{acl_natbib}

\end{document}